\documentclass[11pt]{article}

\usepackage{naacl2021}

\usepackage{times}
\usepackage{latexsym}

\usepackage{microtype}

\usepackage{amsmath,amssymb,amsfonts}
\usepackage{algorithmic}
\usepackage{graphicx}
\usepackage{textcomp}
\usepackage{xcolor}
\usepackage{multirow}
\usepackage{tabularx}
\usepackage{amsmath}
\usepackage[ruled, linesnumbered]{algorithm2e}
\usepackage{xcolor}
\usepackage{multicol}
\usepackage{amsfonts}
\usepackage{subfigure}
\usepackage[toc,page]{appendix}
\usepackage{float}
\usepackage{dblfloatfix}
\usepackage{longtable}
\usepackage{adjustbox}
\usepackage{float, makecell}

\usepackage{array}
\newcolumntype{L}[1]{>{\raggedright\let\newline\\\arraybackslash\hspace{0pt}}m{#1}}
\newcolumntype{C}[1]{>{\centering\let\newline\\\arraybackslash\hspace{0pt}}m{#1}}
\newcolumntype{R}[1]{>{\raggedleft\let\newline\\\arraybackslash\hspace{0pt}}m{#1}}

\usepackage{tikz,xcolor,hyperref}
\usepackage{hyperref}
\usepackage{academicons}
\definecolor{lime}{HTML}{A6CE39}
\DeclareRobustCommand{\orcidicon}{%
	\begin{tikzpicture}
	\draw[lime, fill=lime] (0,0) 
	circle [radius=0.16] 
	node[white] {{\fontfamily{qag}\selectfont \tiny ID}};
	\draw[white, fill=white] (-0.0625,0.095) 
	circle [radius=0.007];
	\end{tikzpicture}
	\hspace{-2mm}
}

\foreach \x in {A, ..., Z}{%
	\expandafter\xdef\csname orcid\x\endcsname{\noexpand\href{https://orcid.org/\csname orcidauthor\x\endcsname}{\noexpand\orcidicon}}
}

\title{Emotion Classification in a Resource Constrained Language Using Transformer-based Approach}

\author{Avishek Das, Omar Sharif, Mohammed Moshiul Hoque \and Iqbal H. Sarker\\
   Department of Computer Science and Engineering \\
    Chittagong University of Engineering \& Technology\\
    Chittagong-4349, Bangladesh \\
  \texttt{u1504018@student.cuet.ac.bd,\{omar.sharif,moshiul\_240,iqbal\}@cuet.ac.bd}\\}

\begin{document}
\maketitle              

\begin{abstract}
Although research on emotion classification has significantly progressed in high-resource languages, it is still infancy for resource-constrained languages like Bengali. However, unavailability of necessary language processing tools and deficiency of benchmark corpora makes the emotion classification task in Bengali more challenging and complicated. This work proposes a transformer-based technique to classify the Bengali text into one of the six basic emotions: anger, fear, disgust, sadness, joy, and surprise.  A Bengali emotion corpus consists of 6243 texts is developed for the classification task. Experimentation carried out using various machine learning (LR, RF, MNB, SVM), deep neural networks (CNN, BiLSTM, CNN+BiLSTM) and transformer (Bangla-BERT, m-BERT, XLM-R) based approaches. Experimental outcomes indicate that XLM-R outdoes all other techniques by achieving the highest weighted $f_1$-score of $69.73\%$ on the test data. The dataset is publicly available at \url{https://github.com/omar-sharif03/NAACL-SRW-2021}.

\end{abstract}

\section{Introduction}
Classification of emotion in the text signifies the task of automatically attributing an emotion category to a textual document selected from a set of predetermined categories. With the growing number of users in virtual platforms generating online contents steadily as a fast-paced, interpreting emotion or sentiment in online content is vital for consumers, enterprises, business leaders, and other parties concerned. Ekman \cite{ekman1993facial} defined six basic emotions: happiness, fear, anger, sadness, surprise, and disgust based on facial features. These primary type of emotions can also be extracted from the text expression \cite{alswaidan2020survey}.

The availability of vast amounts of online data and the advancement of computational processes have accelerated the development of emotion classification research in high-resource languages such as English, Arabic, Chinese, and French \citep{del2020emoevent}. However, there is no notable progress in low resource languages such as Bengali, Tamil and Turkey. 
The proliferation of the Internet and digital technology usage produces enormous textual data in the Bengali language. The analysis of these massive amounts of data to extract underlying emotions is a challenging research issue in the realm of Bengali language processing (BLP). The complexity arises due to various limitations, such as the lack of BLP tools, scarcity of benchmark corpus, complicated language structure, and limited resources. By considering the constraints of emotion classification in the Bengali language, this work aims to contribute to the following:

\begin{itemize}
\item Develop a Bengali emotion corpus consisting of 6243 text documents with manual annotation to classify each text into one of six emotion classes: anger, disgust, fear, joy, sadness, surprise. 
\item Investigate the performance of various ML, DNN and transformer-based approaches on the corpus.
\item Proposed a benchmark system to classify emotion in Bengali text with the experimental validation on the corpus.
\end{itemize}

\section{Related Work}
Substantial research activities have been carried out on emotion analysis in high-resource languages like English, Arabic, and Chinese \citep{alswaidan2020survey}. 
A multi-label with multi-target emotion detection of Arabic tweets accomplished using decision trees, random forest, and KNN, where random forest provided the highest $f_1$-score of 82.6\% \citep{alzu2019multi}. \citet{lai2020fine} proposed a graph convolution network architecture for emotion classification from Chinese microblogs and their proposed system achieved an F-measure of 82.32\%. Recently, few works employed transformer-based model (i.e., BERT) analyse emotion in texts. \citep{huang2019ana,al2020emodet2} used a pre-trained BERT for embedding purpose on top of LSTM/BiLSTM to get an improved \textit{$f_1$}-score of 76.66\% and 74.78\% respectively. 

Although emotion analysis on limited resource languages like Bengali is in the preliminary stage, few studies have already been conducted on emotion analysis using ML and DNN methods. \citet{tripto2018detecting} proposed an LSTM based approach to classify multi-label emotions from Bengali, and English sentences. This system considered only YouTube comments and achieved 59.23\% accuracy. Another work on emotion classification in Bengali text carried out by \citet{azmin2019emotion} concerning three emotional labels (i.e., happy, sadness and anger). They used Multinomial Naïve Bayes, which outperformed other algorithms with an accuracy of 78.6\%. \citet{pal2020anubhuti} developed a logistic regression-based technique to classify four emotions (joy, anger, sorrow, suspense) in Bengali text and achieved 73\% accuracy. \citet{das2009word} conducted a study to identify emotions in Bengali blog texts. Their scheme attained 56.45\% accuracy using the conditional random field. Recent work used SVM to classify six raw emotions on 1200 Bengali texts which obtained 73\% accuracy \cite{ruposh2019computational}.

\section{BEmoC: Bengali Emotion Corpus}
Due to the standard corpus unavailability, we developed a corpus (hereafter called `BEmoC') to classify emotion in Bengali text. The development procedure is adopted from the guidelines stated in \citep{dash2019utility}. 

\subsection{Data Collection and Preprocessing}
Five human crawlers were assigned to accumulate data from various online/offline sources. They manually collected 6700 text documents over three months (September 10, 2020 to December 11, 2020). The crawler accumulated data selectively, i.e., when a crawler finds a text that supports the definition of any of the six emotion classes according to \citet{ekman1993facial}, the content is collected, otherwise ignored. Raw accumulated data needs following pre-processing before the annotation:
\begin{itemize}
\item Removal of non-Bengali words, punctuation, emoticons and duplicate data.
\item Discarding data less than three words to get an unerring emotional context.
\end{itemize}
After pre-processing the corpus holds 6523 text data. The processed texts are eligible for manual annotation. The details of the preprocessing modules found in the link\footnote{\url{https://github.com/omar-sharif03/NAACL-SRW-2021/tree/main/Code\%20Snippets}}.

\subsection{Data Annotation and Quality}
Five postgraduate students working on BLP were assigned for initial annotation. To choose the initial label majority voting technique is applied \cite{magatti2009automatic}. Initial labels were scrutinized by an expert who has several years of research expertise in BLP. The expert corrected the labelling if any initial annotation is done inappropriately. The expert discarded 163 texts with neutral emotion and 117 texts with mixed emotions for the intelligibility of this research. To minimize bias during annotation, the expert finalized the labels through discussions and deliberations with the annotators \cite{10.1007/978-3-030-73696-5_2}. We evaluated the inter-annotator agreement to ensure the quality of the annotation using the coding reliability \citep{krippendorff2011agreement} and Cohen's kappa \citep{cohen1960coefficient} scores. An inter-coder reliability of 93.1\% with Cohen's Kappa score of 0.91 reflects the quality of the corpus.



\subsection{Data Statistics}
The BEmoC contains a total of 6243 text documents after the preprocessing and annotation process. Amount of data inclusion in BEmoC varies with the sources. For example, among online sources, Facebook contributed the highest amount (2796 texts) whereas YouTube (610 texts), blogs (483 texts), and news portals (270 texts) contributed a small amount. Offline sources contributed a total of 2084 texts, including storybooks (680 texts), novels (668 texts), and conversations (736 texts). Data partitioned into train set (4994 texts), validation set (624 texts) and test set (625 texts) to evaluate the models. Table \ref{tab:classwise-tvt} represents the amount of data in each class according to the train-validation-test set.

\begin{table}[!ht]
\begin{center}
\small
\begin{tabular}{C{2cm}| c c c}
\hline
\textbf{Class} & \textbf{Train} & \textbf{Validation} & \textbf{Test} \\ \hline
Anger & 621 & 67 & 71 \\ 
Disgust &  1233 & 155 & 165 \\ 
Fear & 700 & 89 & 83 \\ 
Joy & 908 & 120 & 114 \\ 
Sadness & 942 & 129 & 119 \\ 
Surprise & 590 & 64 & 73 \\ \hline

\end{tabular}
\caption{Number of instances in the train, validation, and test sets}
\label{tab:classwise-tvt}
\end{center}
\end{table}

Since the classifier models learn from the training set instances to obtain more insights, we further analyzed this set. Table \ref{tab:train} shows several statistics of the training set.

\begin{table}[ht]
\begin{center}
\small
\begin{tabular}{C{1.4cm} C{1.2cm} C{1.4cm} C{2cm}}
\hline
\textbf{Class} & \textbf{Total words} & \textbf{Unique words} & \textbf{Avg. words per text } \\ \hline
Anger & 14914 & 5852  & 24.02 \\ 
Disgust &  27192 & 7212  & 22.35 \\ 
Fear & 14766 & 5072  & 21.09 \\ 
Joy & 20885 & 7346  & 23.40 \\ 
Sadness & 22727 & 7398  &24.13 \\ 
Surprise & 13833 & 5675  & 23.45\\ \hline
Total & 114317 & 38555 & -\\
\hline
\end{tabular}
\caption{Statistics of the train set of BEmoC}
\label{tab:train}
\end{center}
\end{table}

\begin{figure}[!b]
\centering
\includegraphics[width=\linewidth]{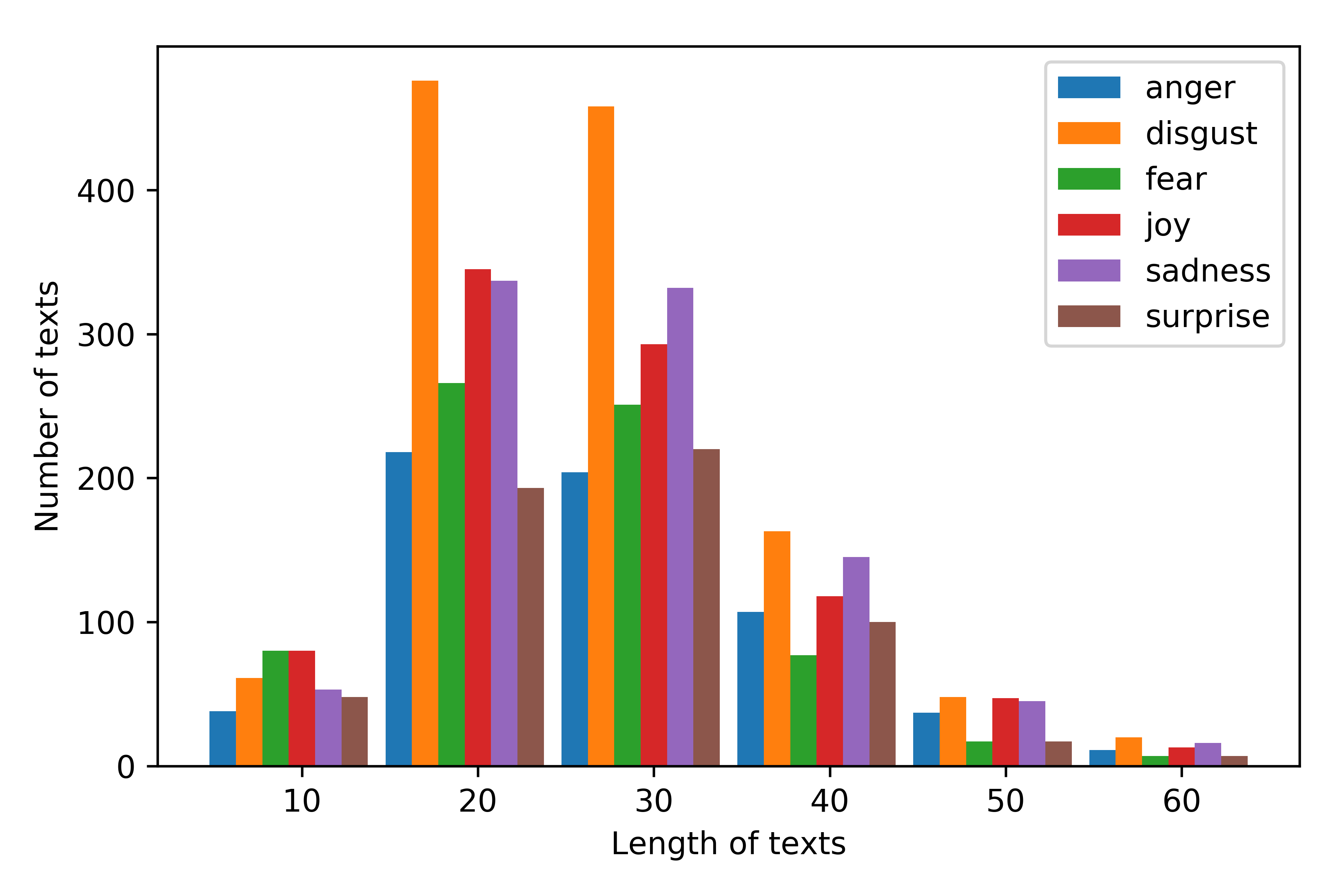}
\caption{Corpus distribution concerning number of texts vs length}
\label{length_plot}
\end{figure}

The \textit{sadness} class contains the most unique words(7398), whereas the \textit{fear} class contains the least(5072). In average all the classes have more than 20 words in each text document. However, a text document in \textit{sadness}  class contained the maximum number of words (107) whereas \textit{fear} class consisting of a minimum number of words (4). Figure \ref{length_plot} represents the number of texts vs the length of texts distribution for each class of the corpus. Investigating this figure revealed that most of the data varied a length between 15 to 35 words.  Interestingly, most of the texts of \textit{Disgust} class have a length less than 30. The \textit{Joy} \& \textit{Sadness} classes seem to have almost similar number of texts in all length distributions.

For quantitative analysis, the \textit{Jaccard} similarity among the classes has been computed. We used 200 most frequent words from each emotion class, and the similarity values are reported in table \ref{jaccerd}. The \textit{Anger-Disgust} and \textit{Joy-Surprise} pairs hold the highest similarity of 0.58 and 0.51, respectively. These scores indicate that more than 50\% frequent words are common in these pair of classes. On the other hand, the \textit{Joy-Fear} pair has the least similarity index, which clarifies that this pair's frequent words are more distinct than other classes. These similarity issues can substantially affect the emotion classification task. Some sample instances of BEmoC are shown in Table 9 (Appendix B).

\begin{table}[!ht]
\small
\begin{center}

\begin{tabular}{c|c|c|c|c|c|c}
\hline
            & \textbf{C1} & \textbf{C2} & \textbf{C3} & \textbf{C4} & \textbf{C5} & \textbf{C6} \\ \hline
\textbf{C1} & 1.00        & 0.58        & 0.40        & 0.43        & 0.45        & 0.47        \\ 
\textbf{C2} & -           & 1.00        & 0.41        & 0.45        & 0.47        & 0.44        \\ 
\textbf{C3} & -           & -           & 1.00        & 0.37        & 0.45        & 0.46        \\ 
\textbf{C4} & -           & -           & -           & 1.00        & 0.47        & 0.51        \\ 
\textbf{C5} & -           & -           & -           & -           & 1.00        & 0.48        \\ \hline
\end{tabular}
\end{center}
\caption{\textit{Jaccard} similarity between the emotion class pairs. Anger (c1), disgust (c2), fear (c3), joy (c4), sadness (c5), surprise (c6).}
\label{jaccerd}
\end{table}



\section{Methodology}
\begin{figure}[!htb]
\centering
\includegraphics[width=0.9\linewidth, height=5.7cm]{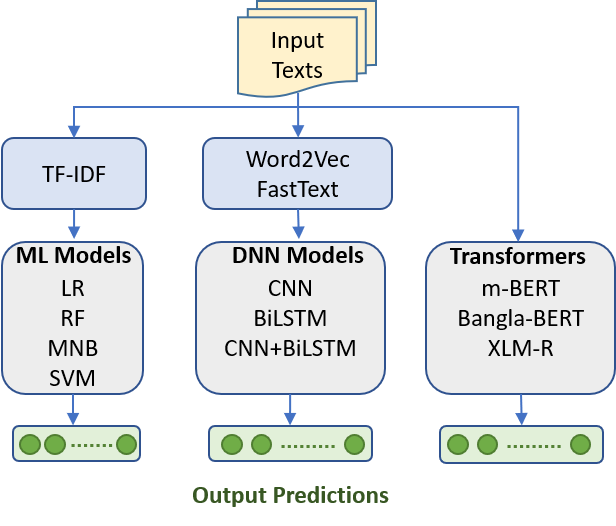}
\caption{Abstract process of emotion classification}
\label{framwork}
\end{figure}

Figure \ref{framwork} shows an abstract view of the used strategies. Various feature extraction techniques such as TF-IDF, Word2Vec, and FastText are used to train ML and DNN models. 
Moreover, we also investigate the Bengali text's emotion classification performance using transformer-based models.
All the models are trained and tuned on the identical dataset. 

\subsection{Feature Extraction}
ML and DNN algorithms are unable to learn from raw texts. Therefore, feature extraction is required to train the classifier models.


\textbf{TF-IDF:} Term frequency-inverse document frequency (TF-IDF) is a statistical measure that determines the importance of a word to a document in a collection of documents. Uni-gram and bi-gram features are extracted from the most frequent 20000 words of the corpus.

\textbf{Word2Vec:} It utilizes neural networks to find semantic similarity of the context of the words in a corpus \cite{mikolov2013efficient}. 
We trained Word2Vec on Skip-Gram with the window size of 7, minimum word count to 4, and the embedding dimension of 100.

\textbf{FastText:} This technique uses subword information to find the semantic relationships \cite{bojanowski2017enriching}. 
We trained FastText on Skip-Gram with character n-grams of length 5, windows size of 5, and embedding dimension of 100.

For both Word2Vec and FastText, there are pre-trained vectors available for the Bengali language trained on generalized Bengali wiki dump data \citep{sarker2021bnlp}. 
We observed that the deep learning models perform well on vectors trained with our developed BEmoC rather than the pre-trained vectors.

\subsection{ML Approaches}
We started an investigation on emotion detection system with ML models. Logistic Regression (LR), Support Vector Machine (SVM), Random Forest (RF) and Multinomial Naive Bayes (MNB) techniques are employed using TF-IDF text vectorizer. For LR \textit{lbfgs}' solver and `\textit{l1}' penalty is chosen and \textit{C} value is set to 1. The same \textit{C} value with `\textit{linear}' kernal is used for SVM. Meanwhile, for RF `\textit{n\_estimators}' is set to 100 and `\textit{alpha=1.0}' is chosen for MNB. A summary of the parameters chosen for ML models are provided in Table \ref{hyper-ml} (Appendix \ref{hyper-appen}). 

\subsection{DNN Approaches}
Variation of deep neural networks (DNN) such as CNN, BiLSTM and a combination of CNN and BiLSTM (CNN+BiLSTM) will investigate the performance of emotion classification task in Bengali. 
To train all the DNN models, `adam’ optimizer with a learning rate of 0.001 and a batch size of 16 is used for 35 epochs. The `sparse\_categorical\_crossentropy’ is selected as the loss function. 

\begin{table*}[!b]
\begin{center}
\small
\begin{tabular}{C{2.4cm}|C{5cm}|C{1.5cm}|C{1.5cm}|C{1.5cm}|C{1.5cm}}
\hline
\textbf{Method}                              & \textbf{Classifier}            & \textbf{Pr(\%)}& \textbf{Re(\%)}  &\textbf{F1(\%)} & \textbf{Acc(\%)} \\ \hline
\multirow{4}{*}{ML models}      
& LR   & 61.07   & 60.64 & 60.75  & 60.64  \\
                                    & RF         & 55.91   & 54.72 & 52.78  & 54.72  \\
                                    & MNB          & 60.23   & 54.08 & 48.67  & 54.08  \\
                                    & SVM                   & 61.12   & 60.10 & 59.54  & 60.00  \\ \hline
\multirow{6}{*}{DNN models} 
& CNN (Word2Vec)         & 53.20   & 52.12 & 51.84  & 52.12  \\
                                    & CNN (FastText)         & 54.54   & 53.45 & 52.52  & 53.48  \\
                                    & BiLSTM (Word2Vec)     & 56.81   & 55.78 & 53.45  & 57.12  \\
                                    & BiLSTM (FastText)     & 57.30   & 58.08 & 56.94  & 58.08  \\
                                    & CNN + BiLSTM (Word2Vec)  & 56.48   & 56.64 & 56.39  & 56.64  \\
                                    & CNN + BiLSTM (FastText) & 55.74   & 55.68 & 55.41  & 55.68  \\ \hline
\multirow{3}{*}{Transformers} & Bangla-BERT           & 62.08   & 62.24 & 61.91  & 62.24  \\
                                    & m-BERT                & 64.62   & 64.64 & 64.39  & 64.63  \\
                                    & XLM-R           & \textbf{70.11}   & \textbf{69.61 }& \textbf{69.73}  & \textbf{69.61}   \\ \hline 
\end{tabular}
\caption{Comparison of various approaches on test set. Here Acc, Pr, Re, F1 denotes accuracy, weighted precision, recall, and \textit{$f_1$}-score}
\label{scores}
\end{center}
\end{table*}

\textbf{CNN:} Convolutional Neural Network (CNN) \cite{lecun2015deep} is tuned over the emotion corpus. The trained weights from the Word2Vec/FastText embeddings are fed to the embedding layer to generate a sequence matrix. The sequence matrix is then passed to the convolution layer having 64 filters of size 7. The convolution layer’s output is max-pooled over time and then transferred to a fully connected layer with 64 neurons. `ReLU’ activation is used in the corresponding layers. Finally, an output layer with softmax activation is used to compute the probability distribution of the classes.

\textbf{BiLSTM:} Bidirectional Long-Short Term Memory (BiLSTM) \cite{hochreiter1997long} is a variation of recurrent neural network (RNN). 
The developed BiLSTM network consists of an Embedding layer similar to CNN, a BiLSTM layer with 32 hidden units, and a fully connected layer having 16 neurons with `ReLU’ activation. An output layer with `softmax’ activation is used.

\textbf{CNN+BiLSTM:} An embedding layer followed by a 1D convolutional layer with 64 filters of size three and a 1D max-pool layer is employed on top of two BiLSTM layers with 64 and 32 units. Outputs of BiLSTM layer fed to an output layer with `softmax’ activation.

Table \ref{hyper} (Appendix \ref{hyper-appen}) illustrates the details of the hyperparameters used in the DNN models.


\subsection{Transformer Models}
We used three transformer models: m-BERT, Bangla-BERT, and XLM-R on BEmoC. In recent years transformer is being used extensively for classification tasks to achieve state-of-the-art results \citep{chen2021transformer}. The models are culled from the Huggingface\footnote{\url{https://huggingface.co/transformers/}} transformers library and fine-tuned on the emotion corpus by using Ktrain\cite{maiya2020ktrain} package.

\textbf{m-BERT:} m-BERT \cite{devlin2019bert} is a transformer model pre-trained over 104 languages with more than 110M parameters. We employed ‘bert-base-multilingual-cased’ model and fine-tuned it on BEmoC with a batch size of 12.

\textbf{Bangla BERT:} Bangla BERT \cite{Sagor_2020} is a pre-trained BERT mask language modelling, trained on a sizeable Bengali corpus. We used the ‘sagorsarker/Bangla-bert-base’ model and fine-tuned to update the pre-trained model fitted for BEmoC. A batch size of 16 is used to provide better results.

\textbf{XLM-R:} XLM-R \cite{liu2019roberta} is a sizeable multilingual language model trained on 100 different languages. 
We implemented the `xlm-Roberta-base’ model on BEmoC with a batch size of 12.

All the transformer models have been trained with 20 epochs with a learning rate of $2e^{-5}$. By using the checkpoint best intermediate model is stored to predict on the test data. Table \ref{hyper-transformers} (Appendix \ref{hyper-appen}) shows a list of parameters used for transformer models.

\section{Results and Analysis}
This section presents a comprehensive performance analysis of various ML, DNN, and transformer-based models to classify Bengali texts emotion. The superiority of the models is determined based on the weighted $f_1$-score. However, the precision ($Pr$), recall ($Re$) and accuracy ($Acc$) metrics also considered. Table \ref{scores} reports the evaluation results of all models.

Among ML approaches, LR achieved the highest (60.75\%) $f_1$-score than RF (52.78\%), MNB (48.67\%) and SVM (59.54\%). LR also performed well in $Pr$, $Re$ and $Acc$ than other ML models. In DNN, BiLSTM with FastText outperformed other approaches concerning all the evaluation parameters. It achieved $f_1$-score of 56.94\%. However, BiLSTM (FastText) achieved about 4\% lower $f_1$-score than the best ML method (i.e., LR).

After employing transformer-based models, it observed a significant increase in all scores. Among transformer-based models, Bangla-BERT achieved the lowest of 61.91\% $f_1$-score. However, this model outperformed the best ML and DNN approaches (56.94\% for BiLSTM (FastText) and 60.75\% for LR). Meanwhile, m-BERT shows almost 3\% improved $f_1$-score (64.39\%) than Bangla-BERT (61.91\%). XLM-R model shows an immense improvement of about 6\% compared to Bangla-BERT and 5\% compared to m-BERT, respectively. It achieved a $f_1$-score of 69.73\% that is the highest among all models.

\subsection{Error Analysis}
\begin{figure}[!ht]
    \begin{center}
    \includegraphics[width=\linewidth]{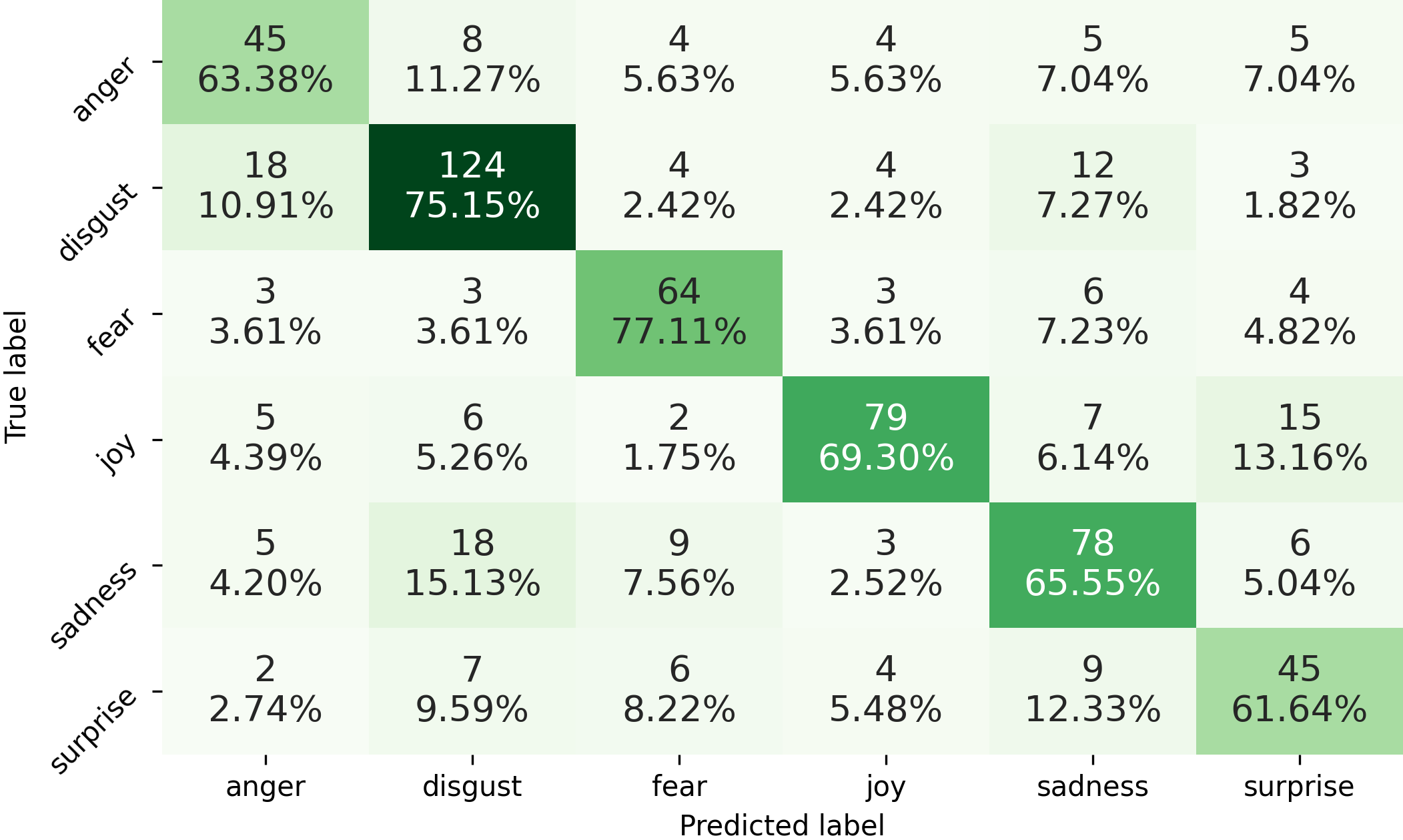}
\caption{Confusion matrix of XLM-R model}
\label{conf_mat}
\end{center}
\end{figure}

It is evident from Table \ref{scores} that XLM-R is the best performing model to classify emotion from Bengali texts. A detailed error analysis is performed using the confusion matrix. Figure \ref{conf_mat} illustrates a class-wise proportion of the number of predicted labels.
It is observed from the matrix that few data classified wrongly. For example, 8 instances among 71 of the \textit{anger} class predicted as \textit{disgust}. In \textit{fear} class, 6 data out of 83 mistakenly classified as \textit{sadness}. In the \textit{sadness} class, misclassification ratio is the higest (15.13\%). That means, 18 data out of 119  in \textit{sadness} class misclassified as \textit{disgust}. Moreover, among 73 data in \textit{surprise} class 9 are predicted as \textit{sadness}. The error analysis reveals that \textit{fear} class achieved the highest rate of correct classification (77.15\%) while \textit{surprise} gained the lowest (61.64\%).

The possible reason for incorrect predictions might be the class imbalance nature of the corpus. However, the high value of \textit{Jaccard similarity} (Table \ref{jaccerd}) also reveals some interesting points. Few words are used multi-purposely in multiple classes. For instance, hate words can be used to express both \textit{anger} and \textit{disgust} feelings. Moreover, emotion classification is highly subjective, depends on the individual’s perception, and people may contemplate a sentence in many ways \citep{LEDOUX201867}. Thus, by developing a balanced dataset with diverse data, incorrect predictions might be reduced to some extent.

\subsection{Comparison with Recent Works}
The analysis of results revealed that XLM-R is the best model to classify emotion in Bengali texts. Thus, we compare the performance of XLM-R with the existing techniques to assess the effectiveness. We implemented previous methods \cite{tripto2018detecting,azmin2019emotion,pal2020anubhuti,ruposh2019computational} on BEmoC and reported outcomes in $f_1$-score. Table \ref{compare_work} shows a summary of the comparison.  
\begin{table}[!h]
\begin{center}
\small
\begin{tabular}{L{5.3cm}|C{1.4cm}}
\hline
\textbf{Methods} &\textbf{F1(\%)} \\ \hline

Word2Vec + LSTM \cite{tripto2018detecting} &  53.54 \\ 
TF-IDF + MNB \cite{azmin2019emotion} &  48.67\\
TF-IDF + LR \cite{pal2020anubhuti} &  60.75 \\ 
BOW + SVM \cite{ruposh2019computational} &  59.17 \\ 
XLM-R (\textbf{Proposed}) &   \textbf{69.73} \\ \hline
\end{tabular}
\caption{Performance comparison. Here F1 denotes weighted \textit{$f_1$}-score.}
\label{compare_work}
\end{center}
\end{table}
The results show that XLM-R outperformed the past techniques with achieving the highest \textit{$f_1$}-score (69.73\%).

\section{Conclusion}
This paper investigated various ML, DNN and transformer-based techniques to classify the emotion in Bengali texts. Due to the scarcity of benchmark corpus, we developed a corpus (i.e., BEmoC) containing 6243 Bengali texts labelled with six basic classes. Co-hen’s Kappa score of 0.91 reflects the quality of the corpus. Performance analysis on BEmoC illustrated that XLM-R, a transformer model provided a superior result among all the methods. Specifically, XLM-R achieved the highest $f_1$-score of 69.61\% which indicates the improvement of 8.97\% (than ML) and 11.53\% (than DNN). Although XLM-R exhibited the most elevated scores, other technique (such as the ensemble of the transformer models) can also investigate enhancing performance. Additional categories (such as love, hate, and stress) can also include generalization. Moreover, transformer-based models can also investigate extending the corpus, including text with sarcasm or irony, text with comparison and mixed-emotion.

\section*{Acknowledgements}
We sincerely acknowledge the anonymous reviewers and pre-submission mentor for their insightful suggestions, which help improve the work. This work was supported by the Directorate of Research \& Extension, CUET.

\bibliographystyle{acl_natbib}
\bibliography{bibliography}

\clearpage
\appendix
\section*{Appendices}
\label{Appendix}
\section{Model Hyperparameters}
\label{hyper-appen}


\begin{table}[!h]
\centering
\begin{tabular}{L{1.8cm}|C{5cm}} 
\hline
\textbf{ Classifier } & \textbf{ Parameters }                                        \\ 
\hline
LR & optimizer = ‘lbfgs’, max\_iter = 400, penalty = ‘l1’, C=1  \\ 
\hline
SVM & kernel=‘linear’, random\_state = 0, $\gamma$=‘scale’,  tol=‘0.001’      \\ 
\hline
RF & criterion=‘gini’, n\_estimators = 100                          \\ 
\hline
MNB & $\alpha$ = 1.0, fit\_prior = true, class\_prior = none,                  \\
\hline
\end{tabular}
\caption{Optimized parameters for ML models}
\label{hyper-ml}
\end{table}


\begin{table}[!ht]
\centering
\begin{tabular}{c|C{4.5cm}|C{1.8cm}|C{1.8cm}|C{2cm}}
\hline

\textbf{Hyperparameters} &
  \textbf{Hyperparameter Space} &
  \textbf{CNN}&
 \textbf{BiLSTM}&
 \textbf{CNN + BiLSTM}\\ \hline
 
Filter Size &
3,5,7,9 &
7&
-&
3
\\ \hline
Pooling type & ‘max’, ‘average’ & ‘max’ & - &  ‘max’
\\ \hline
Embedding  Dimension &
  30, 35, 50, 70, 90, 100, 150,  200, 250, 300 &
  100 &
  100 &
  100 \\ \hline
Number of Units &
  16,   32, 64, 128, 256 &
  64 &
  32&
  64,64,32\\ \hline
Neurons in Dense Layer &
  16,   32, 64, 128, 256 &
  64 &
  16 &
  -\\ \hline
  Batch Size &
   16,   32, 64, 128, 256 &
   16 &16 & 16
   \\ \hline
Activation Function &
 `relu', `tanh', `softplus',  `sigmoid'  &
  `relu' &
  `relu' &
  `relu' \\ \hline
Optimizer &
 `RMSprop', `Adam', `SGD',   `Adamax'  & 
 `Adam'  & 
 `Adam'  & 
 `Adam' \\ \hline
 Learning Rate &
  0.5, 0.1, 0.05, 0.01, 0.005,  0.001, 0.0005, 0.0001 &
  0.001 &
  0.001 &
  0.001 \\ \hline
   
\end{tabular}

\caption{Hyperparameters for DNN methods}
\label{hyper}
\small
\end{table}

\begin{table}[!h]
\centering
\begin{tabular}{c|c} 
\hline
\textbf{ Hyperparameter } & \textbf{ Value }  \\ 
\hline
Fit method                & ‘auto\_fit’       \\ 
\hline
Learning rate             & 2e-5              \\ 
\hline
Epochs                    & 20                \\ 
\hline
Batch size                & 12,16                \\ 
\hline
Max sequence length       & 70        \\
\hline
\end{tabular}
\caption{Optimized hyperparameters for transformers models}
\label{hyper-transformers}
\end{table}

\clearpage

\begin{figure*}[!htb]
\centering
\includegraphics[width=\linewidth]{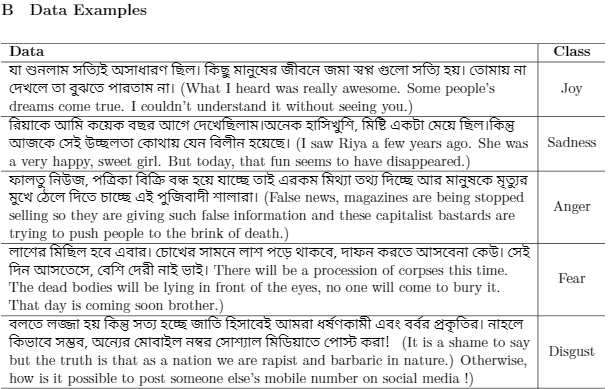}
\vspace{-4cm}

\end{figure*}
\begin{figure*}[!htb]
\centering
\includegraphics[width=0.99\linewidth]{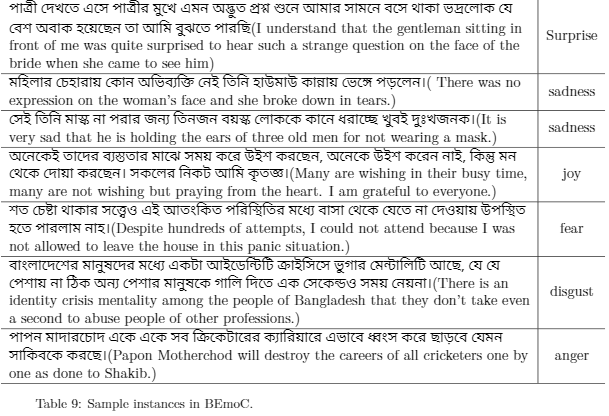}

\end{figure*}


\end{document}